\documentclass[runningheads]{llncs}
\usepackage[T1]{fontenc}

\usepackage{graphicx}
\usepackage{amsmath} 
\usepackage{amssymb}
\usepackage{amsfonts} 
\usepackage{mathrsfs}
\usepackage{booktabs}
\begin{document}
\pagenumbering{gobble}

\title{Does simple trump complex? Comparing strategies for adversarial robustness in DNNs} 
\titlerunning{Does simple trump complex?}
%
\author{William Brooks\inst{1,2,4}\orcidID{0009-0001-4373-8350} \and
Marelie H Davel\inst{1,2,3}\orcidID{0000-0003-3103-5858} \and
Coenraad Mouton\inst{1,2}\orcidID{0000-0001-8610-2478}}
\authorrunning{Brooks et al.}
%
\institute{Faculty of Engineering,
North-West University, South Africa \\
\and Centre for Artificial Intelligence Research, South Africa
\and National Institute for Theoretical and Computational Sciences, South Africa
\and South African National Space Agency \\
\email{williambrooks499@gmail.com}\\
\url{http://engineering.nwu.ac.za/must}}
\maketitle              
\begin{abstract}
Deep Neural Networks (DNNs) have shown substantial success in various applications but remain vulnerable to adversarial attacks. This study aims to identify and isolate the components of two different adversarial training techniques that contribute most to increased adversarial robustness, particularly through the lens of margins in the input space -- the minimal distance between data points and decision boundaries.
Specifically, we compare two methods that maximize margins: a simple approach which modifies the loss function to increase an approximation of the margin, and a more complex state-of-the-art method (Dynamics-Aware Robust Training) which builds upon this approach. Using a VGG-16 model as our base, we systematically isolate and evaluate individual components from these methods to determine their relative impact on adversarial robustness. We assess the effect of each component on the model's performance under various adversarial attacks, including AutoAttack and Projected Gradient Descent (PGD). Our analysis on the CIFAR-10 dataset reveals which elements most effectively enhance adversarial robustness, providing insights for designing more robust DNNs. 

\keywords{Deep Neural Networks \and Adversarial robustness \and Margins in input space \and Dynamics-Aware Robust Training \and Large margin loss \and Adversarial attacks}
\end{abstract}
\section{Introduction}
Deep neural networks (DNNs) have transformed numerous fields and have achieved remarkable success in tasks ranging from image classification to natural language processing. Despite their widespread adoption, DNNs are inherently vulnerable to adversarial perturbations -- small perturbations to the input data that lead a deep learning model to misclassify the input, often with high confidence \cite{szegedy2014intriguingpropertiesneuralnetworks}. This vulnerability poses challenges where reliability and security of machine learning models are paramount. 

The exact reason for the susceptibility of DNNs to these perturbations remains unknown and is an ongoing area of research. Several methods have been developed to generate adversarial examples and evaluate the robustness of DNNs \cite{croce2020minimallydistortedadversarialexamples,croce2020reliable,madry2018towards,moosavidezfooli2016deepfoolsimpleaccuratemethod}. Popular attack algorithms, such as Projected Gradient Descent (PGD)~\cite{madry2018towards}, have become standard tools for testing the robustness of machine learning models. These attack algorithms aim to exploit the vulnerability of neural networks to adversarial examples and current research aims to improve the robustness of DNNs, by building defenses to such perturbations \cite{bai2024mixednutstrainingfreeaccuracyrobustnessbalance,bartoldson2024adversarial,rebuffi2021fixingdataaugmentationimprove,wang2023betterdiffusionmodelsimprove}. 

Popular defences against adversarial attacks, such as Adversarial Training (AT)~\cite{goodfellow2015explainingharnessingadversarialexamples,madry2018towards}, aim to enhance the robustness of these models by explicitly exposing them to adversarial examples during training. Despite being effective, this approach is computationally expensive as it requires generating adversarial perturbations during training and does not guarantee robustness against all types of attacks. Margin-based techniques, while often similarly computationally expensive, have been explored as an alternative due to their ability to offer a better robustness-accuracy trade-off compared to methods like PGD adversarial training \cite{elsayed2018large,xu2023exploring}. 
Margin-based techniques focus on increasing the margin -- the minimum distance between a sample and its decision boundary -- in the input space. Large margins in the input space have been shown to be correlated with improved adversarial robustness of DNNs \cite{Ding2020MMA}.

The work by Xu et al. \cite{xu2023exploring} and Elsayed et al. \cite{elsayed2018large} primarily focused on increasing margins, utilizing innovative approaches to margin maximization (MM) of DNNs. Elsayed et al. introduced a novel loss function based on the first-order Taylor approximation of the margin, which maximizes the margin at any layer, demonstrating improvements in adversarial robustness. Xu et al. built on this concept by identifying the vulnerability of data points with smaller margins and proposing Dynamics-Aware Robust Training (DyART), a technique that prioritizes increasing these smaller margins in the input space. DyART achieved near state-of-the-art performance on benchmarks such as RobustBench~\cite{croce2021robustbenchstandardizedadversarialrobustness} improving on recognized baselines such as AT~\cite{madry2018towards}, TRADES~\cite{zhang2019theoreticallyprincipledtradeoffrobustness}, MMA~\cite{Ding2020MMA}, GAIRAT \cite{zhang2021geometryaware}, MAIL~\cite{wang2022probabilisticmarginsinstancereweighting} and AWP~\cite{wu2020adversarialweightperturbationhelps} under various perturbation bounds. 

Both the approaches by Elsayed et al.~\cite{elsayed2018large} and Xu et al.~\cite{xu2023exploring} focus on MM and introduce loss functions for increasing margins in the input space to improve the adversarial robustness of DNNs. Despite the progress made by these margin-based techniques, the techniques themselves are complex and it is not clear which elements most contribute 
to enhancing adversarial robustness. To address this question,  we conduct a detailed analysis of the adversarial robustness of models trained using the techniques proposed by Xu et al. and Elsayed et al. We use AutoAttack~\cite{croce2020reliable}, a state-of-the-art evaluation method, to systematically compare the performance of these methods and identify the key factors contributing to their robustness. 
The primary contributions of this study are as follows: 
\begin{enumerate}
    \item An evaluation of the adversarial robustness of models trained using Xu et al. and Elsayed et al.'s techniques,
    \item Identification of the impactful components of these methods in enhancing adversarial robustness,
    \item Insights into the aspects which matter when constructing adversarially robust classifiers.
\end{enumerate} 
The remainder of this paper is organized as follows: Section \ref{sec:background} provides an overview of adversarial training, margin-based techniques, and the work by Xu et al.~\cite{xu2023exploring} and Elsayed et al.~\cite{elsayed2018large}. 
Section \ref{sec:Approach} describes the approach followed to disentangle the effects of the underlying elements of each technique and describes the individual components analysed. 
Section \ref{sec:experiments} describes the experimental setup, 
while the results of the analysis are presented in Section \ref{sec:Results}.

\section{Background}\label{sec:background}
In this section, we provide background information on adversarial training and the two main methods we consider. First (Section~\ref{sec:at_and_evaluation}) we provide a brief introduction to adversarial training in general, and how the robustness of a model is typically evaluated. Following this, we describe how margins are defined and discuss the commonalities of adversarial training techniques that maximize margins (Section~\ref{sec:margin_maximization}). 
Finally, we provide an overview of the MM methods of Elsayed et al. and Xu et al. in Sections \ref{sec:elsayed} and \ref{sec:dyart}, respectively.

\subsection{Adversarial Training and Evaluation}
\label{sec:at_and_evaluation}

Adversarial examples pose a considerable challenge to the robustness of deep neural networks (DNNs). The vulnerability of DNNs to these adversarial examples has sparked significant research into methods that can mitigate their effect. Although AT is the standard approach for enhancing the robustness of classifiers, it often results in a decrease in clean accuracy (accuracy when considering standard samples, not adversarially manipulated ones), a phenomenon known as the accuracy-robustness trade-off~\cite{tsipras2018robustness}. Therefore, current research aims to develop robust training methods able to provide the best possible robustness to adversarial attacks with as little possible drop in performance on clean examples. This is an active area of research and many methods have been proposed~\cite{Ding2020MMA,xu2023exploring,zhang2019theoreticallyprincipledtradeoffrobustness}. 


Conversely, the goal of adversarial attack algorithms are to fool a classifier by slightly perturbing clean examples. This has led to a research arms race where stronger adversarial attacks are developed to fool robust training methods, examples of which are Projected Gradient Descent (PGD)~\cite{madry2018towards}, Fast Adaptive Boundary (FAB)~\cite{croce2020minimallydistortedadversarialexamples}, DeepFool~\cite{moosavidezfooli2016deepfoolsimpleaccuratemethod}, Fast Gradient Sign Method (FGSM)~\cite{goodfellow2015explainingharnessingadversarialexamples}, and the Carlini\&Wagner (C\&W) attack~\cite{carlini2017towards}. These attacks can be used to generate adversarial examples, but can also be used to evaluate the robustness of DNNs to these attacks. 
However, training a model using these attack-specific adversarial perturbed examples can lead to overfitting to specific types of attacks. Although a model trained with AT might be robust against PGD attacks, for example, it might not generalize well to other, potentially unknown, attacks~\cite{ilyas2019adversarialexamplesbugsfeatures}. 
It is therefore important to measure the robustness of models using current state-of-the-art (SOTA) approaches that combine multiple attacks.

AutoAttack~\cite{croce2020reliable} has become the standard benchmark for evaluating the adversarial robustness of modern DNNs. AutoAttack consists of an ensemble of four attacks. These four attacks include two variations of the standard PGD attack developed by Madry et al.~\cite{madry2018towards}: an enhanced version of PGD known as Auto-PGD with cross-entropy (APGD-CE) and the targeted Auto-PGD (APGD-t) \cite{croce2020reliable}. Additionally it includes the targeted FAB~\cite{croce2020minimallydistortedadversarialexamples} attack, and the black-box Square attack~\cite{andriushchenko2020square}.  Croce et al. established a framework, RobustBench \cite{croce2021robustbenchstandardizedadversarialrobustness}, that utilizes AutoAttack with the intention of being a standardized benchmark to measure the robust accuracy of DNNs, and to compare findings and results.

\subsection{Margin Maximization}
\label{sec:margin_maximization}

When employing adversarial training, it is necessary to distinguish between techniques that directly utilize adversarial examples -- training with both clean samples and adversarial examples -- and those that focus on MM. Prominent MM approaches include TRADES \cite{zhang2019theoreticallyprincipledtradeoffrobustness}, the work by Elsayed et al.~\cite{elsayed2018large}, DyART~\cite{xu2023exploring}, and MMA~\cite{Ding2020MMA}.

These approaches leverage the concept of a margin, denoted as $R$ from here onwards, defined as the minimum distance of a sample to its nearest decision boundary. Consider a deep classifier $f$ parameterized by $\theta$, with inputs $x \in X$ and corresponding labels $y \in Y = \{1, 2, \dots, n\}$, that generates a prediction score to classify the input vector $x$ to class $y$. We can define the prediction of the classifier $f_\theta$ as the class $y$ corresponding to the maximum output logit $z_\theta^y(x)$ given by $f_\theta(x) = \arg\max_{y \in Y} z_\theta^y(x)$. 
To define the decision boundary, we introduce $\phi$, which represents the logit margin -- the difference between the logit value for the true class $y$ and the highest logit value among all other classes $y' \neq y$:
\begin{equation}\label{eq:logit_margin}
 \phi_\theta(x) = z_\theta^y(x) - \max_{y' \neq y} z_\theta^{y'}(x)
\end{equation}
A sample $x$ is deemed to be correctly classified when $\phi_\theta^y(x) >0$ and incorrectly classified otherwise. 
The decision boundary exists where there is a tie, that is, $\phi_\theta^y(x) = 0$. 
We can then define the margin, $R_\theta$, as the distance of a sample $x$ to the nearest sample $\hat{x}$ on the decision boundary:
%
\begin{equation}\label{eq:min_margin_def}
 R_\theta(x) = \min_{\hat{x}} \|\hat{x} - x\|_p \quad \text{s.t.} \quad \phi_\theta^y(\hat{x}) = 0
\end{equation}
Here $\|.\|_p$ is some chosen $l_p$ norm. We typically use the $l_\infty$ norm as is convention~\cite{madry2018towards,xu2023exploring}. 

The core idea of MM is to utilize a loss function that combines traditional cross-entropy loss ($\mathcal{L}_{\text{CE}}$) with a MM loss term ($\mathcal{L}_{\text{MM}}$). This hybrid loss function, as shown in Equation \ref{eq:basic_mm_loss_func}, is designed to increase margins in the input space, thereby improving the model's adversarial robustness during training.
\begin{equation}\label{eq:basic_mm_loss_func}
\mathcal{L} = \mathcal{L}_{\text{CE}}(y,\hat{y}) + \lambda \cdot \mathcal{L}_{\text{MM}}(\gamma, R)
\end{equation}
Here 
$y$ is the true label of a sample and $\hat{y}$ the model's corresponding prediction. The $\mathcal{L}_{\text{MM}}$ term typically includes a user-defined hyperparameter, $\gamma$, which represents the minimum margin that should be maintained, making the model less susceptible to perturbations of an input.
In addition, a hyperparameter $\lambda$ is incorporated to weigh the margin maximization component of the loss function. 

We will primarily focus on the MM loss functions proposed by Elsayed et al. and Xu et al., respectively. 
Both methods utilize hybrid loss functions that aim to maximize margins in the input space whilst minimizing the loss with respect to the inputs for correct classification \cite{elsayed2018large,xu2023exploring}. 

\subsection{Large Margin Loss Approach}
\label{sec:elsayed}

The $L_{MM}$ function introduced by Elsayed et al., for a batch $\mathcal{B}$ of size $N$, is defined as:
\begin{equation}\label{eq:large_margin_loss_function}
\small
\mathcal{L}_{MM}(\gamma, R) = \frac{1}{N}\sum_{i \in \mathcal{B}} \mathscr{A}_{y' \neq y} \max \left\{0, \gamma - R_\theta(x_i) \right\}
\end{equation}
Here $\mathscr{A}_{y' \neq y}$ is some aggregation operator, and they experiment with choosing this as either the \textit{max} operator or the \textit{sum} operator. 
For each training sample $x$ and corresponding label $y$, the individual loss is computed for each class where $y' \neq y$. The margin of a sample $x$ is therefore calculated to the decision boundary of every other class, $y' \neq y$. 
These individual losses are then aggregated using the max or sum operator.

Intuitively, this loss function aims to ensure that all samples have at least a margin of $\gamma$ (the minimum margin) to the nearest decision boundary. Additionally, it penalizes samples with smaller margins more than those with larger margins, and also accommodates both correctly and incorrectly classified samples. Consider a correctly classified sample with a margin smaller than the minimum margin. The margin loss for this sample will then be equal to $\gamma - R$. However, if the samples has a margin larger than the minimum margin, it will simply be assigned a loss of zero.  On the other hand, if a sample is incorrectly classified, the margin ($R$) will be negative and the loss for that sample will be adjusted to account for both the distance the decision boundary must move for it to be correctly classified and have a positive margin of at least $\gamma$.

The behavior of this loss function is perhaps made more clear if one considers how Elsayed et al. measures the margin for each sample. While some authors measure the exact distance to an adversarial example to measure the margin, Elsayed et al. opt to rather approximate this distance. This greatly reduces the computational cost, as calculating adversarial examples for each batch of training data can be highly expensive. Specifically, they utilize a first order Taylor approximation, as shown in Equation \ref{eq:elsayed_margin}, to approximate the margin $R$. 

\begin{equation}\label{eq:elsayed_margin}
R_\theta(x)=\frac{z_\theta^y(x) - z_\theta^{y'}(x)}{\left\|\nabla_{\boldsymbol{x}} z_\theta^y(x)-\nabla_{\boldsymbol{x}} z_\theta^{y'}(x)\right\|_q},
\end{equation}
The numerator in Equation~\ref{eq:elsayed_margin} is given by the difference between the output logits of the true class, $y$, and some other class, $y'$ (i.e. the same as Equation \ref{eq:logit_margin}). The denominator computes the difference between the gradients \(\nabla_{\boldsymbol{x}}\) of $z_\theta^y(x)$ and $z_\theta^{y'}(x)$ with regard to to the input $\boldsymbol{x}$, and then calculates the norm \(\left\|.\right\|_q\). Calculating the gradient terms requires calculating the Jacobian matrix which gives the partial derivatives of the logits with respect to the input $x$. Here, $\|.\|_q$ denotes the dual norm of $\|.\|_p$ where $\frac{1}{q} + \frac{1}{p} = 1$. Specifically, when $p = \infty$, then $q = 1$.
Furthermore, it should be clear that if the sample is incorrectly classified, the numerator will be negative, and the resulting margin will also be a negative value.
%
%
Since there is a derivative in the MM loss function, it requires calculating the Hessian (second-order gradient) when calculating the gradient of the loss w.r.t. to the model parameters for gradient descent, which is computationally too expensive. They therefore treat the denominator in Equation \ref{eq:elsayed_margin} as a constant during back-propagation. 

Finally, Elsayed et al. also experiment with extending this loss function to the margin of each sample as measured at the hidden layers. Specifically, they replace the input $x$ with its hidden representation $h_l$ at a given layer $l$. Their goal is to enforce a large margin based on the data representation at each specific layer. We do not consider the maximization of margins measured at the hidden layers in this work.

\subsection{DyART}
\label{sec:dyart}

Similar to the approach by Elsayed et al., Xu et al. designed a MM loss function. They identified an issue with other robust training methods where  increasing the margin for one datapoint might decrease the margin of another data point. They called this `conflicting dynamics'. To mitigate this issue, they introduced DyART, an MM loss function designed to prioritize increasing smaller margins \cite{xu2023exploring}. This ensures that all samples have a margin of at least $\gamma$.
Their MM function is defined as follows:
\begin{equation}\label{eq:xu_exp_margin_loss}
    MM(R)=\left\{\begin{array}{cc}
\frac{1}{\alpha} \exp (-\alpha R), & R<\gamma \\
0, & \text { otherwise }
\end{array}\right.
\end{equation}
This function assigns an exponential loss to samples with a margin smaller than $\gamma$; otherwise, the loss is zero. The hyperparameter, $\alpha$, smooths this exponential function. However, unlike the MM function of Elsayed et al., the loss function is only applied when $\phi_\theta^y(x) > 0$, meaning incorrectly classified samples are ignored. 
Since most samples are still incorrectly classified at the start of training, a burn-in period or training warmup is recommended, where the model is trained naturally using cross-entropy for a certain number of epochs before applying the MM loss function. The loss for a training batch $\mathcal{B}$ of size $N$, with correctly classified samples $\mathcal{B}_{\theta}^+$ of size $m$, is defined as: 
\begin{equation}
    \mathcal{L}_{MM}(R) = \frac{1}{m} \sum_{i\in\mathcal{B}_\theta^+} MM(R_\theta(x_i))
\end{equation}
Xu et al. also use a more accurate estimation of the margin for each sample than Elsayed et al. Instead of relying on the first-order Taylor approximation, DyART utilizes FAB \cite{croce2020minimallydistortedadversarialexamples} to find the closest boundary point $\hat{x}$ from sample $x$. The margin is then given by $||x - \hat{x}||_\infty$. FAB iteratively applies the Taylor approximation to find a point on the decision boundary, and is much more computationally expensive. In addition, if the margin is calculated to each class it needs to calculate the Jacobian at each iteration between the true class and every other class. Therefore, finding the closest point on the decision boundary becomes computationally prohibitive, especially for datasets with a large number of classes, such as CIFAR-100. FAB also requires multiple iterations to produce high quality adversarial examples. 

To lessen this computational burden, Xu et al. do not calculate the margin to each non-label class for each samples. Instead, they introduced an alternative method for calculating the margin, known as the soft margin, $R_\theta^{\text{soft}}(x)$, which is the minimum distance of sample to a point on the `soft decision boundary'. The soft decision boundary is an approximation of the exact, or `true', decision boundary.  
More specifically, the soft decision boundary is given by the points where $\phi_\theta^y(x ; \beta) = 0$ for a hyperparameter $\beta$, and $\phi_\theta^y(x ; \beta)$ is defined as:
\begin{equation}\label{eq:xu_soft_des_boundary}
    \phi_\theta^y(x ; \beta)=z_\theta^y(x)-\frac{1}{\beta} \log \sum_{y^{\prime} \neq y} \exp \left(\beta z_\theta^{y^{\prime}}(x)\right)
\end{equation}

Therefore, instead of calculating the distance to the exact decision boundary, which requires evaluating the logits for all possible pairs of classes, as defined in Equation \ref{eq:logit_margin}, the Log-Sum-Exp function gives a weighted average which emphasizes the largest logits without needing to compare each logit. This reduces the number of operations required as it avoids the need to compute the margin to every possible pair of logits. Moreover, this soft decision boundary is controlled by a hyperparameter, $\beta$, and the higher this value, the closer the soft decision boundary represents the exact decision boundary. 


Finally, recall that Elsayed et al. are unable to calculate the appropriate gradient of $L_{MM}$ w.r.t. the model parameters during training due to computational constraints. Xu et al. encounter the same problem, since the margin calculation (recall Equation~\ref{eq:min_margin_def}), is within itself an optimization problem. 
Inspired by MMA \cite{Ding2020MMA}, Xu et al. derive a closed-form expression that allows efficiently calculating the gradient of $L_{MM}$ w.r.t. the model parameters. 
This approach allows for the model parameters to be updated based on a more accurate representation of how the margin changes, rather than approximating it as done by Elsayed et al.\cite{elsayed2018large} and Ding et al.\cite{Ding2020MMA}.

\section{Approach}\label{sec:Approach}
As should be clear from Section 2, the DyART adversarial training technique can be considered a more complicated version of Elsayed's that introduces several improvements. Our approach is to progressively introduce each of these improvements individually to Elsayed's loss function to isolate and identify design choices that lead to improved adversarial robustness.

In this section we elaborate on the different models we train and the expected behavior of each design choice. We provide more extensive implementation and hyperparameter details in the following section. We first describe our baseline models, in Section~\ref{sec:approach_baseline}. Following this, we describe the different elements we introduce to Elsayed's method in Section~\ref{sec:approach_additions}. Finally, in Section~\ref{sec:approach_evaluating} we describe how we evaluate each model's performance.

\subsection{Baseline models}
\label{sec:approach_baseline}

We start by implementing three baseline models using the CIFAR-10 dataset, which we partition into 45,000 training samples and 5,000 validation samples: 1) a standard model trained  with vanilla cross-entropy loss, 2) a model trained using Elsayed's MM loss function, and 3) a model trained with Xu's DyART MM loss function. This allows us to compare each change we make to the original methods, as well as compare to a `normally' trained model.

We also make some slight changes to the original implementations to suit our needs.  In terms of Elsayed's method, we opt to only approximate the margin to the highest predicted non-label class, instead of calculating the margin for every class and using an aggregation operator. This reduces the computational burden significantly. Additionally, as mentioned in Section~\ref{sec:elsayed}, Elsayed et al. experimented with increasing the margins at the hidden layers as well. We only apply his approach in the input space in order to make it comparable with the DyART method.

For the DyART method of Xu et al. we also make some slight changes. They implemented stochastic weight averaging \cite{izmailov2019averagingweightsleadswider} for improved performance. We omit this as to better isolate the effect of each loss function individually. In addition to this, they also train on an additional $10$M synthetically generated images to further improve their robustness results. We omit this for the same reason.

\subsection{Additions to Elsayed}
\label{sec:approach_additions}

In this section we elaborate on each of the elements we add to Elsayed's MM method.

\subsubsection{Elsayed’s MM with Burn-in.}
Our approach begins with the Burn-in Phase, where we enhance the standard MM approach by Elsayed et al. by incorporating an initial burn-in period. During this phase, the model undergoes natural training for several epochs using standard cross-entropy loss before any MM techniques are applied. This helps the model establish a solid baseline with relatively high clean accuracy and a greater proportion of correctly classified samples. To optimize this phase, we also need to determine the best set of hyperparameters to maximize the number of correctly classified samples before transitioning to the margin-maximization process. The reasoning behind this is that a strong initial baseline may facilitate the MM process, potentially leading to enhanced adversarial performance. 
\subsubsection{Elsayed’s MM with Burn-in and Exponential Loss.}
Following the burn-in phase, we experiment with the Exponential Loss Function, as defined in Equation \ref{eq:xu_exp_margin_loss}. This function is designed to operate on correctly classified samples, which necessitates the burn-in period, and encourages the model to maintain a margin of at least $\gamma$. Because this loss function is exponential, it penalizes samples with smaller margins more heavily than those with larger margins. Incorporating this into Elsayed's loss function might be beneficial in improving the adversarial robustness due to the way smaller margins are managed and penalized. 
\subsubsection{Elsayed’s MM with FAB.}
Next, we replace Elsayed’s Taylor approximation Equation \ref{eq:elsayed_margin} with FAB, aiming to achieve a more precise margin approximation. FAB accurately calculates the closest boundary points, offering a significant improvement over the original first-order Taylor approximation method. This however also requires adapting FAB to function similarly to our adapted Elsayed et al, to only consider the highest predicted class instead of all the classes. This provides some insight into the importance of a much more accurate approximation of the margin, meaning if it is worth the added computational cost and if this is crucial for increasing the adversarial robustness of DNNs. 
\subsubsection{Elsayed’s MM with FAB and Soft Boundary.}
Finally, we integrate the Soft Boundary technique with our adapted FAB to balance the trade-off between computational cost and the accuracy of margin estimation. Comparing the results of utilizing FAB on its own versus with the soft boundary will shed more light on the importance of a more accurate approximation of the margin and whether reducing the compute by utilizing their soft boundary approach can be beneficial. 

By systematically testing the listed components in a carefully considered order: beginning with the Burn-in Phase, followed by the Exponential Loss Function, then Improved Margin Approximation with FAB, and finally the integration of the Soft Boundary, we aim to develop a better understanding of how each element contributes to the overall robustness of the model.

\subsection{Evaluating Robustness}
\label{sec:approach_evaluating}

For each model, we evaluate its adversarial robustness in several ways. We rely on existing adversarial attack algorithms, and select three specific attacks and two key use cases. First, we use standard PGD as used by Madry et al.~\cite{madry2018towards} to generate adversarial examples for each epoch during training using the validation set. These examples serve as an additional validation set, where we calculate the accuracy and robust loss of our model. This approach helps mitigate robust overfitting -- where prolonged training leads to increased robust loss despite decreasing (clean) training loss \cite{rice2020overfittingadversariallyrobustdeep} -- by applying early stopping on the adversarial validation set.

The second use case involves evaluating the adversarial robustness of our final models. We rely on two different attack algorithms: APGD-CE \cite{croce2020reliable} -- an improved version of the standard PGD attack \cite{madry2018towards}, and finally AutoAttack \cite{croce2020reliable}. While AutoAttack serves as a strong benchmark, APGD-CE allows us to gain a more in-depth understanding of how our models respond to different gradient-based attacks. This combination provides sufficient information to evaluate the adversarial robustness of our models and gain insights into the contributions of the different components.

\section{Experimental Setup}\label{sec:experiments}

In this section, we elaborate on our implementation and hyperparameter details for the experiments outlined in the previous section. First, in Section~\ref{sec:architecture_and_standard_training}, we describe the dataset and architecture we consider, along with the standard training hyperparameters. Following this, in Section~\ref{sec:sweeps}, we elaborate on how hyperparameters were tuned for each model individually.


\subsection{Architecture and Standard Training Parameters}
\label{sec:architecture_and_standard_training}

In this section we explain our standard training setup, including the dataset, architecture, and training procedure we use. 
We use the CIFAR10~\cite{krizhevsky2009learning} for all of our experiments. This is a popular dataset for investigating adversarial robustness~\cite{bartoldson2024adversarial,croce2021robustbenchstandardizedadversarialrobustness,elsayed2018large,xu2023exploring}. This is because even though the dataset is considered relatively simple, building a robust CIFAR10 classifier is still an open problem~\cite{bartoldson2024adversarial}. Furthermore, we specifically focus on robustness to $l_{\infty}$ attacks with a maximum perturbation bound of $\delta = \frac{8}{255}$, which is also a popular evaluation case to investigate on CIFAR10~\cite{croce2021robustbenchstandardizedadversarialrobustness}.

In terms of model architecture, Xu et al. and Elsayed et al. both conducted experiments using larger CNN-based architectures, such as Wide Residual Networks and ResNet-18s. However, these are very large models that are expensive to train. Instead, to accommodate our computational budget, we rely on the smaller but well known VGG-16 architecture~\cite{simonyan2015deepconvolutionalnetworkslargescale} for all experiments.  

We train our models using stochastic gradient descent with a momentum of 0.9, employing cosine annealing for the learning rate without restarts and a minimum learning rate decay of 0.001 \cite{LoshchilovH16a}. Cosine annealing allows us to set an initial learning rate and define a target learning rate to which it should decay or increase. This approach enables us to maintain constant learning rates during training or to implement learning rate schedules with specific patterns, such as increasing or decreasing the learning rates from some initial learning rate to some specified target value. 
The models are trained for 200 epochs, and we use batch normalization on all layers. Xu et al. also experimented with using group normalization \cite{wu2018group}, but we find that batch normalization offers better performance.

For model selection and to mitigate robust overfitting \cite{rice2020overfittingadversariallyrobustdeep}, we created an additional adversarial validation set consisting of 1,024 samples generated from the original validation set using a 20-step PGD attack \cite{madry2018towards} with a perturbation bound of $\delta = \frac{8}{255}$. Early stopping is performed based on the model's performance on this adversarial validation set, and the model with the highest adversarial validation accuracy was selected. 

\subsection{Hyperparameter Optimization}
\label{sec:sweeps}

In this section, we expand on how hyperparameters were chosen for each model. We list each model and describe the hyperparameter optimization process. From each sweep, we select two models: the model with the highest clean validation accuracy, and the model with the highest validation APGD-CE robust accuracy. We report on the results of these hyperparameter optimizations in the following section.

\subsubsection{Baseline VGG16.}

We performed extensive hyperparameter sweeps to identify the best settings for our VGG-16 model. Initially, we explored various batch sizes (64, 128, 256, 512), learning rates (0.1, 0.01, 0.001), and weight decay values (0.001, 0.0005, 0.0002) to determine the optimal configuration. The model with the highest validation accuracy used a batch size of 128, a learning rate of 0.1, and a weight decay of 0.0005. This model achieves a very respectable $89.24\%$ accuracy on the CIFAR10 test set. We also use these hyperparameters as a starting point for further experiments. 

\subsubsection{DyART.}

To find optimal hyperparameters for the VGG16 model using the DyART training method, we conduct the hyperparameter sweep in two separate phases. First, we conducted a sweep to optimize the burn-in period. This phase involved experimenting with several learning rates ($0.1$, $0.01$, $0.001$) and batch sizes ($64$, $128$, $256$, $512$, $1024$). Since the DyART training method is very computationally expensive (due to the expensive margin-measuring method), we limited the burn-in phase to $25$ epochs of standard training followed by only $3$ epochs using the DyART loss function. For the hyperparameters specific to the DyART MM loss, we use the exact same hyperparameters as used by Xu et al.~\cite{xu2023exploring}, which are $\alpha =3$, $\gamma=\frac{16}{255}$, $\lambda=1000$, $\beta=5$, and $\delta=\frac{8}{255}$.
For the second phase, we selected the top $10$ models' hyperparameters based on their adversarial validation accuracy from the initial sweep.  We then proceeded to retrain these models for 25 epochs of natural training followed by 200 epochs with the DyART loss function. 


\subsubsection{Elsayed's Margin Maximization Loss.} For the MM loss function of Elsayed et al., we conducted a full sweep to determine the optimal learning rate ($0.1$, $0.01$, $0.001$) and batch size ($128$, $256$, $512$). In addition, it was necessary to also search over values of $\lambda$ ($10$, $15$, $20$, $25$), as the scale of the margin loss is different than when using Xu's method and therefore requires a different weighting with cross-entropy~\footnote{We could also not rely on the $\lambda$ values chosen by Elsayed et al., as they included the margins of hidden layers which also changes the scale.}
We found that a batch size of $128$, a learning rate of $0.1$ and $\lambda=25$ yielded the best adversarial validation accuracy.


\subsubsection{Additions to Elsayed}

For each component that we add to Elsayed's loss function (as described in Section~\ref{sec:Approach}), we use the hyperparameters that yielded the highest robust validation accuracy from the hyperparameter sweep described above, for the vanilla implementation of Elsayed's margin maximzation loss. However, we also search over different learning rates for each addition to ensure performance.
This is with the exception of the addition of Xu's exponential margin loss, which necessitates changing $\lambda$ as it changes the scale of the margin loss. In the following section we present results using both $\lambda=25$ (as used for Elsayed's method) and $\lambda=1000$ (as used for Xu's method).

\section{Results}\label{sec:Results}
This section presents the outcome of the experiments designed to optimize adversarial robustness for the VGG-16 model using the approaches detailed in Section \ref{sec:Approach} and setup defined in Section \ref{sec:experiments}. As outlined in Section~\ref{sec:sweeps}, we conducted extensive hyperparameter sweeps across various configurations, including different learning rates, batch sizes, and regularization parameters, to identify the optimal settings for both DyART and MM loss functions. The results of these sweeps are summarized in Table~\ref{tab:Learning_rate_sweeps_on_robustness_components}.%
\begin{table}[htbp]
\centering
\caption{
Performance of VGG-16 models evaluated under a perturbation bound of 8/255. 'Clean Accuracy' is reported for both validation and test sets, while 'Robust Accuracy' includes results from AutoAttack and APGD-CE. For each configuration — DyART and Elsayed's MM function — we report the results for models with the highest clean accuracy and the highest APGD accuracy on the test set.} 
\label{tab:Learning_rate_sweeps_on_robustness_components}
\small
\begin{tabular}{lcccccc}
\toprule
\textbf{Model} & \multicolumn{2}{c}{\textbf{Robust Accuracy}} & \multicolumn{2}{c}{\textbf{Clean Accuracy}} \\
\cmidrule(lr){2-3} \cmidrule(lr){4-5}
 & \textbf{AutoAttack} & \textbf{APGD-CE} & \textbf{Val} & \textbf{Test} \\
\midrule
\multicolumn{6}{c}{\textbf{VGG-16 Baseline}} \\
\midrule
Baseline & 0 & 0 & 90.08 & 89.24\\
\midrule
\multicolumn{6}{c}{\textbf{Elsayed's MM function}} \\
\midrule
Best Clean Accuracy & 0 & 13.17 & 88.64 & 87.62\\
Best APGD Accuracy & 0 & 14.29 & 88.70 & 87.35\\
\midrule
\multicolumn{6}{c}{\textbf{Elsayed's MM with Burn-in}} \\
\midrule
Best Clean Accuracy & 0 & 13.30 & 88.12 & 87.00\\
Best APGD Accuracy & 0 & 27.64 & 87.34 & 86.44\\
\midrule
\multicolumn{6}{c}{\textbf{Elsayed's MM with Burn-in and Exponential Loss $\lambda=25$}} \\
\midrule
Best Clean Accuracy & 0 & 9.22 & 87.84 & 87.07\\
Best APGD Accuracy & 0 & 26.78 & 86.30 & 85.15\\
\midrule
\multicolumn{6}{c}{\textbf{Elsayed's MM with Burn-in and Exponential Loss $\lambda=1000$}} \\
\midrule
Best Clean Accuracy & 0 & 47.38 & 88.32 & 87.51\\
Best APGD Accuracy & 0.13 & 49.26 & 84.58 & 83.60\\
\midrule
\multicolumn{6}{c}{\textbf{Elsayed's MM with FAB}} \\
\midrule
Best Clean Accuracy & 0 & 4.46 & 88.54 & 87.47 \\
Best APGD Accuracy & 0 & 5.40 & 85.98 & 85.24 \\
\midrule
\multicolumn{6}{c}{\textbf{Elsayed's MM with FAB and Soft Boundary}} \\
\midrule
Best Clean Accuracy & 0 & 2.80 & 87.38 & 86.86 \\
Best APGD Accuracy & 0 & 25.70 & 86.32 & 85.26 \\
\midrule
\multicolumn{6}{c}{\textbf{DyART}} \\
\midrule
Best Clean Accuracy & 0.01 & 18.77 & 85.60 & 84.89\\
Best APGD Accuracy & 33.24 & 33.52 & 74.72 & 72.81\\
\bottomrule
\end{tabular}
\end{table}
We first discuss the observations for each result in Table~\ref{tab:Learning_rate_sweeps_on_robustness_components} individually, before considering the results as a whole in the following section. 
\subsubsection{VGG-16 Baseline:} The VGG-16 baseline shows no robustness under strong adversarial attacks, with an APGD-CE and AutoAttack accuracy of 0\%. This is to be expected, as the model has not undergone any training to withstand adversarial attacks.
\subsubsection{Elsayed's MM function:} Introducing Elsayed's MM function drastically improves the model's robustness, with the best APGD accuracy model achieving $14.29\%$ accuracy on APGD-CE, while maintaining a relatively high clean accuracy of $87.35\%$. This initial comparison shows the effectiveness of Elsayed's MM function in significantly enhancing adversarial robustness compared to the baseline without a substantial drop in clean accuracy. However, we do not observe any gains in accuracy when using AutoAttack.
\subsubsection{Elsayed’s MM with Burn-in:} The burn-in phase to Elsayed's MM function further improves robustness. The model with the best APGD-CE accuracy reaches $27.64\%$, almost doubling the robustness compared to the MM-only approach. The clean accuracy sees a slight decrease, but this trade-off can be justified by the considerable gain in robustness. This indicates that first allowing the model to learn from clean examples before attempting to maximize the margins is the better strategy for achieving robustness. That said, against the stronger adversary of AutoAttack, we still observe an accuracy of $0\%$.
\subsubsection{Elsayed’s MM with Burn-in and Exponential Loss:} The exponential loss with $\lambda=25$ leads to mixed results. While the best APGD- accuracy model still performs well with an APGD-CE of $26.78\%$, this is slightly lower than with burn-in alone. However, when $\lambda$ is increased to $1000$, there is a significant boost in robustness, with the APGD-CE jumping to $49.26\%$, a substantial improvement over previous configurations. This illustrates that the $\lambda$ parameter must be carefully calibrated to strike an optimal balance between standard cross-entropy loss and the MM loss. We also find a $0.13\%$ robust accuracy on AutoAttack using the larger value for $\lambda$, albeit at the cost of a noticeable reduction in clean accuracy, highlighting the increased trade-off as robustness improves. This points to the notion that the exponential function is effective (to some extent) in mitigating the `conflicting dynamics' issue identified by Xu et al. (recall Section~\ref{sec:dyart}). 
\subsubsection{Elsayed’s MM with FAB:} Utilizing FAB to estimate the margin in Elsayed's MM function results in a decrease in APGD-CE, with the best model achieving only $5.40\%$ accuracy. Even when a soft boundary is added, the maximum APGD-CE accuracy achieved is $25.70\%$, which is lower than the robustness obtained with the exponential loss approach (which uses the first-order Taylor approximation). This indicates that FAB, while an alternative, might not be as effective in this context when compared to the Taylor approximation. This is especially striking if one considers that FAB is much more computationally expensive.

\subsubsection{DyART:} DyART shows different behavior compared to variations of Elsayed's method. While it has a APGD-CE of $33.52\%$, this does not surpass the APGD-CE score of the `Elsayed's MM with Burn-in and Exponential Loss $\lambda=1000$' method. However, we do observe a very respectable $33.24\%$ accuracy when considering the stronger AutoAttack adversary, which is a substantial improvement on all of the other methods. We suspect that this stems from DyART's more accurate calculation of the margin gradients. That said, we note that DyART also has a more aggressive trade-off between robustness and clean accuracy. 

\subsection{Discussion}

When considering the results of the previous section as a whole, we note that all configurations with Elsayed's MM function surpass the VGG-16 baseline in terms of robustness. This clearly demonstrates the effectiveness of these techniques in defending against adversarial attacks, but also demonstrates the trade-off with clean accuracy. Furthermore, these models unfortunately do not achieve any robustness improvements when evaluated using AutoAttack.

It is also evident that some of the additions made by Xu et al. are more effective at improving robustness than others. More specifically, we note that incorporating the exponential loss function achieved a $34.97\%$ increase in robust accuracy evaluated using APGD-CE over the standard approach by Elsayed et al. Likewise, we find that incorporating a burn-in period of standard training is also very beneficial to improving robust accuracy. Conversely, and quite surprisingly, we find that a more accurate and more computationally expensive approximation of the margin with FAB did not improve the adversarial robustness of these models.

Finally, when comparing the variations of Elsayed et al's. method to DyART, we find that the addition of the margin gradients leads to a very large increase in AutoAttack robust accuracy, but simultaneously, a larger trade-off between the robustness and clean accuracy of the models.


\section{Conclusions}
\label{sec:discussion}
While both Elsayed et al.’s and Xu et al.’s margin maximization techniques improve the adversarial robustness of deep neural networks in the input space, the specific elements contributing to this improvement was unknown. Our study aimed to address this by providing a detailed overview of the two techniques and isolating the individual elements of the techniques, evaluating their performance using current SOTA adversarial attack algorithms. 

We have identified certain key elements that greatly contribute to adversarial robustness in DNNs, as well as certain design choices that do not have a large effect. More specifically, we find that:
\begin{enumerate}
    \item A period of standard training before introducing margin maximization (known as a burn-in period) is a simple yet effective trick to increase adversarial robustness. 
    \item Penalizing samples with smaller margins more is very beneficial, such as through the use of an exponential loss function. 
    \item Using a more accurate estimation of the margin does not necessarily lead to improved robustness.
    \item An accurate calculation of the gradient of the margin with regard to the model parameters greatly improves adversarial robustness.
\end{enumerate}


In conclusion, our study provides insights into which elements positively contribute towards designing more robust image classification models. 
While the studied techniques initially seem complex, it has been shown that they can be decomposed into a number of fairly simple elements, each of which can be analysed individually. 
From the observed results, it is expected that future developments that target more accurate and efficient gradient estimations will be more beneficial than techniques aiming to produce more accurate estimations of the margin itself.

\subsubsection{Acknowledgements}

This work is based on research supported in part by the National Research Foundation of South Africa (Ref Numbers PSTD23042296065, RA211019646111).

%
%
%
\pagebreak
\bibliographystyle{splncs04}
\bibliography{mybibliography}

\end{document}